\documentclass[letterpaper, 10 pt, conference]{ieeeconf}  
\usepackage{arydshln}

\IEEEoverridecommandlockouts                              

\usepackage{graphics} 
\usepackage{epsfig} 
\usepackage{mathptmx} 
\usepackage{times} 
\usepackage{amsmath} 
\usepackage{amssymb}  
\usepackage{cite}
\usepackage{cite}
\usepackage{amsmath,amssymb,amsfonts}
\usepackage{algorithmic}
\usepackage{graphicx}
\usepackage{textcomp}
\usepackage{xcolor}
\usepackage{times} 
\usepackage{amsmath} 
\usepackage{amssymb}  
\usepackage{multirow}
\usepackage{multicol}
\usepackage{url}
\usepackage{afterpage}
\usepackage{stfloats}
\usepackage{cite}
\usepackage{arydshln}
\usepackage{epsfig} 
\usepackage{mathptmx}
\usepackage{subcaption}
\usepackage{tablefootnote}
\let\labelindent\relax
\usepackage{enumitem}
\usepackage{colortbl}  
\usepackage{xcolor}    
\definecolor{lightblue}{rgb}{0.85, 0.92, 1.0}  
\usepackage{arydshln}

\title{\LARGE \bf
ELMAR: Enhancing LiDAR Detection with 4D Radar Motion Awareness and Cross-modal Uncertainty
}

\author{Xiangyuan Peng   \hspace{3mm}   Miao Tang    \hspace{3mm}  Huawei Sun \hspace{3mm}  Bierzynski Kay \hspace{3mm} Lorenzo Servadei    \hspace{3mm}  Robert Wille
\thanks{Xiangyuan Peng and Huawei Sun are with Infineon Technologies AG and the Technical University of Munich, Germany,
        {\tt\small Xiangyuan.Peng@infineon.com}}%
\thanks{Miao Tang is with the China University of Geosciences, Wuhan, China}%
\thanks{Bierzynski Kay is with Infineon Technologies AG, Germany}%
\thanks{Lorenzo Servadei and Robert Wille are with the Technical University of Munich, Germany}%
}

\begin{document}

\maketitle
\thispagestyle{empty}
\pagestyle{empty}


\begin{abstract}
LiDAR and 4D radar are widely used in autonomous driving and robotics.
While LiDAR provides rich spatial information, 4D radar offers velocity measurement and remains robust under adverse conditions. As a result, increasing studies have focused on the 4D radar-LiDAR fusion method to enhance the perception.
However, the misalignment between different modalities is often overlooked.
To address this challenge and leverage the strengths of both modalities, we propose a LiDAR detection framework enhanced by 4D radar motion status and cross-modal uncertainty. The object movement information from 4D radar is first captured using a Dynamic Motion-Aware Encoding module during feature extraction to enhance 4D radar predictions. Subsequently, the instance-wise uncertainties of bounding boxes are estimated to mitigate the cross-modal misalignment and refine the final LiDAR predictions. 
Extensive experiments on the View-of-Delft (VoD) dataset highlight the effectiveness of our method, achieving state-of-the-art performance with the mAP of 74.89\% in the entire area and 88.70\% within the driving corridor while maintaining a real-time inference speed of 30.02 FPS.

\end{abstract}

\section{Introduction}

Perception plays an important role in autonomous driving and robotics, providing essential information for subsequent path planning and decision-making \cite{song2024robustness}. A key task of perception is object detection, which can accurately identify various road users, including cars, pedestrians, and cyclists, to reduce the risk of collisions and traffic accidents. 
To achieve more reliable autonomous driving systems, different sensors, including cameras, LiDAR, and radar, are employed to enhance detection performance \cite{fan20244d,shariff2024event,li2020deep}.

Cameras can capture rich semantic information through RGB images but lack depth perception. In contrast, LiDAR sensors provide dense 3D point clouds, making them particularly valuable for 3D object detection. Despite the advantage, LiDAR's performance degrades under adverse environments \cite{appiah2024object}. Additionally, LiDAR is not sensitive to the motion of objects and has a limited sensing range due to its wavelength constraints \cite{appiah2024object}.
As an alternative, 4D radar has attracted increasing attention \cite{fan20244d} due to its longer detection range and robustness under adverse environments compared to LiDAR \cite{harlow2024new}. Besides, 4D radar can generate 3D point clouds by processing received radar signals, which have a similar structure to LiDAR point clouds. Moreover, 4D radar is able to measure Doppler velocity, providing valuable dynamic information. However, 4D radar point clouds are significantly sparser than LiDAR point clouds.

\begin{figure}[t]
    \centering
    \includegraphics[width=1\linewidth]{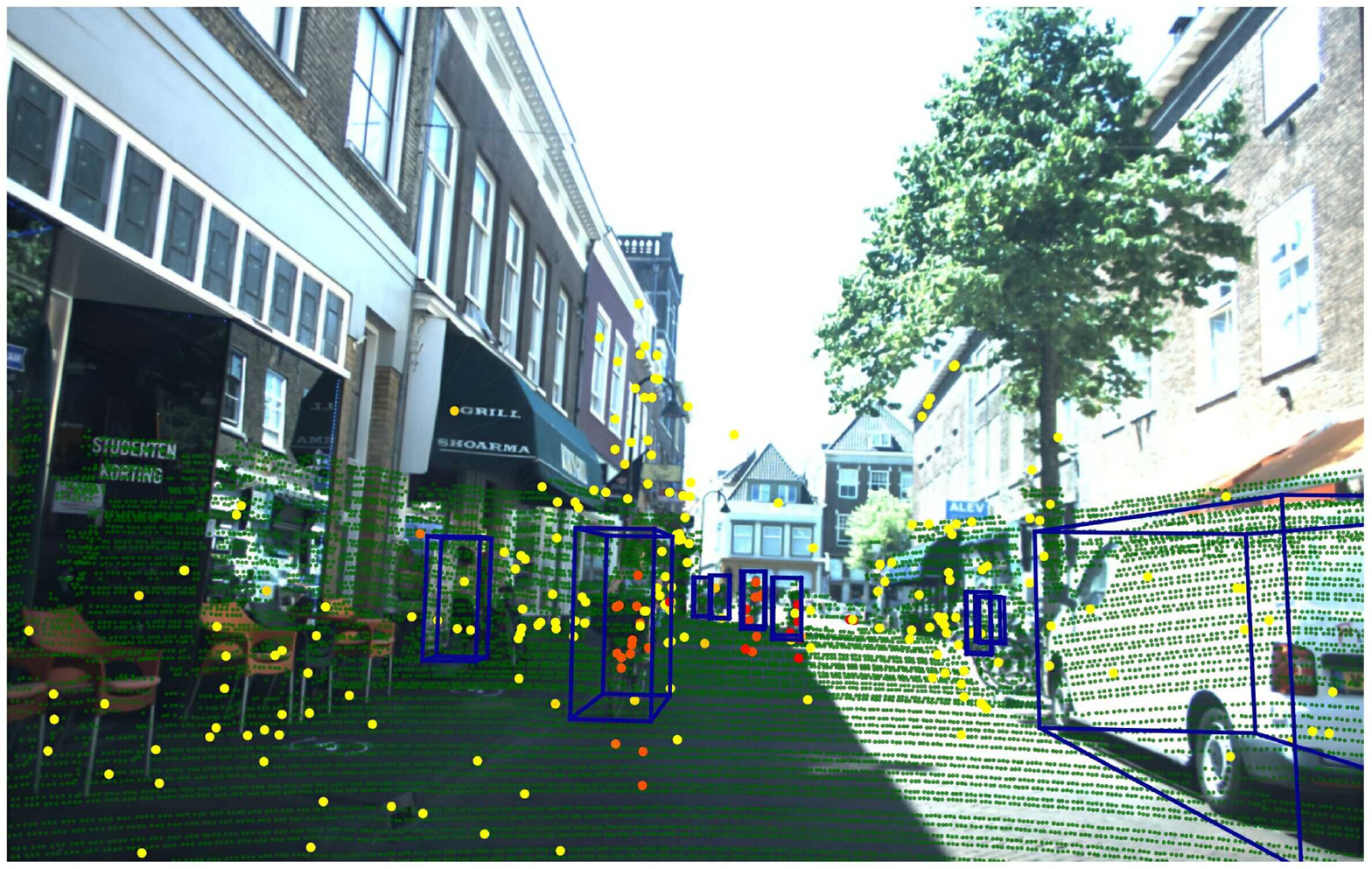}
    \caption{Visualization of the VoD dataset \cite{palffy2022multi}. LiDAR point clouds are marked in green, while 4D radar point clouds are color-coded from yellow to red, where yellow represents low absolute velocity and red indicates high absolute velocity.}
    \label{fig: motivation}
    \vspace{-3mm}
\end{figure}

Therefore, leveraging radar data to enhance LiDAR-based perception is used to improve the accuracy and robustness of 3D object detection in autonomous driving \cite{wang2022interfusion,wang2022multi,deng2024robust}. However, most existing methods didn't explore the objects' motion status, which is crucial for safe perception and decision-making. Traditionally, the object's motion is often estimated by sequential multi-frame LiDAR scans, which leads to the long-tail problem and high computation cost by accumulation \cite{chen2024joint,zeng2024mambamos,10611400}. To mitigate these issues, some studies incorporate radar-measured velocities to introduce dynamic information into LiDAR-based perception.  
For instance, MutualForce \cite{peng2025mutualforce} fuses radar velocity features with LiDAR spatial features at the pillar level. Although radar points provide Doppler velocity measurements, directly applying them and setting fixed thresholds for objects' motion estimation can lead to inaccuracies due to multi-path effects and noise \cite{10161152}.  
As shown in Fig. \ref{fig: motivation}, some radar points outside the bounding boxes of moving objects also have nonzero velocities. Incorporating these velocities can introduce errors in bounding box predictions.

Moreover, the misalignment between 4D radar and LiDAR can degrade the performance of multi-modal detection \cite{huang2024l4dr}. In multi-modal methods, 4D radar is primarily used to enhance LiDAR’s robustness \cite{huang2024l4dr,chae2024towards,wang2023bi}. However, the significant sparsity of 4D radar point clouds can potentially compromise the LiDAR detection performance.

To address these challenges, we propose a novel LiDAR detection framework, ELMAR, enhanced by 4D radar motion status and cross-modal uncertainty. Our approach introduces a 4D radar Dynamic Motion-Aware Encoding (DMAE) module that extracts high-quality dynamic information, improving the sensing of moving objects. Furthermore, we incorporate a Cross-Modal Uncertainty Alignment (X-UA) mechanism that refines the final LiDAR predictions. By aligning the predicted bounding box distributions of both modalities, our framework effectively mitigates the misalignment issues and enhances detection robustness.
The main contributions are as follows:
\begin{itemize}  
    \item A novel single-frame 4D radar-based motion-aware module is designed to implicitly encode the accurate objects' dynamic information.  
    \item A cross-modal uncertainty alignment module is introduced to explicitly reduce the instance-wise misalignment of predictions from different modalities.
    \item Extensive experiments are conducted on the View-of-Delft (VoD) dataset \cite{palffy2022multi}, achieving the state-of-the-art performance.  
\end{itemize}

\section{Related Work}

\subsection{3D Object Detection}
Many 3D object detection methods are based on LiDAR, which can be categorized into voxel-based \cite{lang2019pointpillars,jin2024swiftpillars,chen2023voxelnext}, point-based \cite{yang20203dssd,song2023psns,zhang2022not}, and hybrid approaches \cite{shi2020pv,shi2023pv,jiang2024dpa}. However, LiDAR’s robustness is limited in adverse environments \cite{wallace2020full}. 
On the other hand, 4D radar-based methods implement 3D object detection by leveraging radar's unique characteristics. SMURF \cite{10274127} employs Kernel Density Estimation to extract multi-representational features, while MUFAN \cite{yan2023mvfan} and MUFASA \cite{peng2024mufasa} utilize both cylindrical and bird’s-eye view for improved spatial awareness. However, radar point clouds remain significantly sparse, resulting in a performance drop \cite{jiang2024sparseinteraction}. Therefore, single-modality approaches face notable challenges due to their own drawbacks.

To address the limitations, integrating 4D radar and LiDAR data provides a balanced solution \cite{chae2024towards}. L4DR \cite{huang2024l4dr} utilizes 4D radar data to denoise LiDAR point clouds during preprocessing. The distinct characteristics, such as Radar Cross Section and velocity from radar data, have been leveraged to guide the fusion process in \cite{peng2025mutualforce}. Furthermore, attention mechanisms have been incorporated into pipelines to refine feature integration \cite{wang2022multi}.  
Despite the advancements in 4D radar-LiDAR methods, most existing methods use the directly measured velocity of individual points while neglecting object-level motion states, which limits the ability to distinguish small but moving objects, such as pedestrians.

                           
\subsection{Uncertainty Estimation in Perception}
The uncertainty of the detected objects is a critical factor for ensuring road safety in autonomous driving. Uncertainty estimations are mainly categorized into four strategies.
Ensemble strategy assesses uncertainty by combining the results from multiple deterministic models during the testing process \cite{zheng2021rectifying}. Test-time augmentation strategy applies input transformations during testing and estimates uncertainty based on multiple outputs \cite{wang2024augmenting}. Besides, some approaches design extra uncertainty estimation modules on top of the baseline models \cite{nandy2020towards,lee2020gradients}. However, this strategy can compromise the original model’s performance. Bayesian-based strategy leverages probabilistic neural networks to calculate the variance across multiple forward passes \cite{mobiny2021dropconnect}.

These uncertainty estimation strategies have been widely applied across various tasks. PaSCO \cite{cao2024pasco} estimates voxel- and instance-level uncertainties for scene completion, while internal and external uncertainty estimation have been employed for tracking in \cite{10647838}. Some methods address uncertainty in labels caused by occlusions or annotation errors \cite{liang2023spsnet,zhang2023glenet}. Additionally, uncertainty estimation has also been integrated into LiDAR-camera fusion \cite{lou2023uncertainty,cho2024cocoon}. However, due to the density gap between 4D radar and LiDAR point clouds, the application of uncertainty estimation to align predictions between 4D radar and LiDAR is still lacking.

\section{Proposed Method}

\subsection{Overall Structure}

\begin{figure*}[t]
    \centering
    \includegraphics[width=1\textwidth]{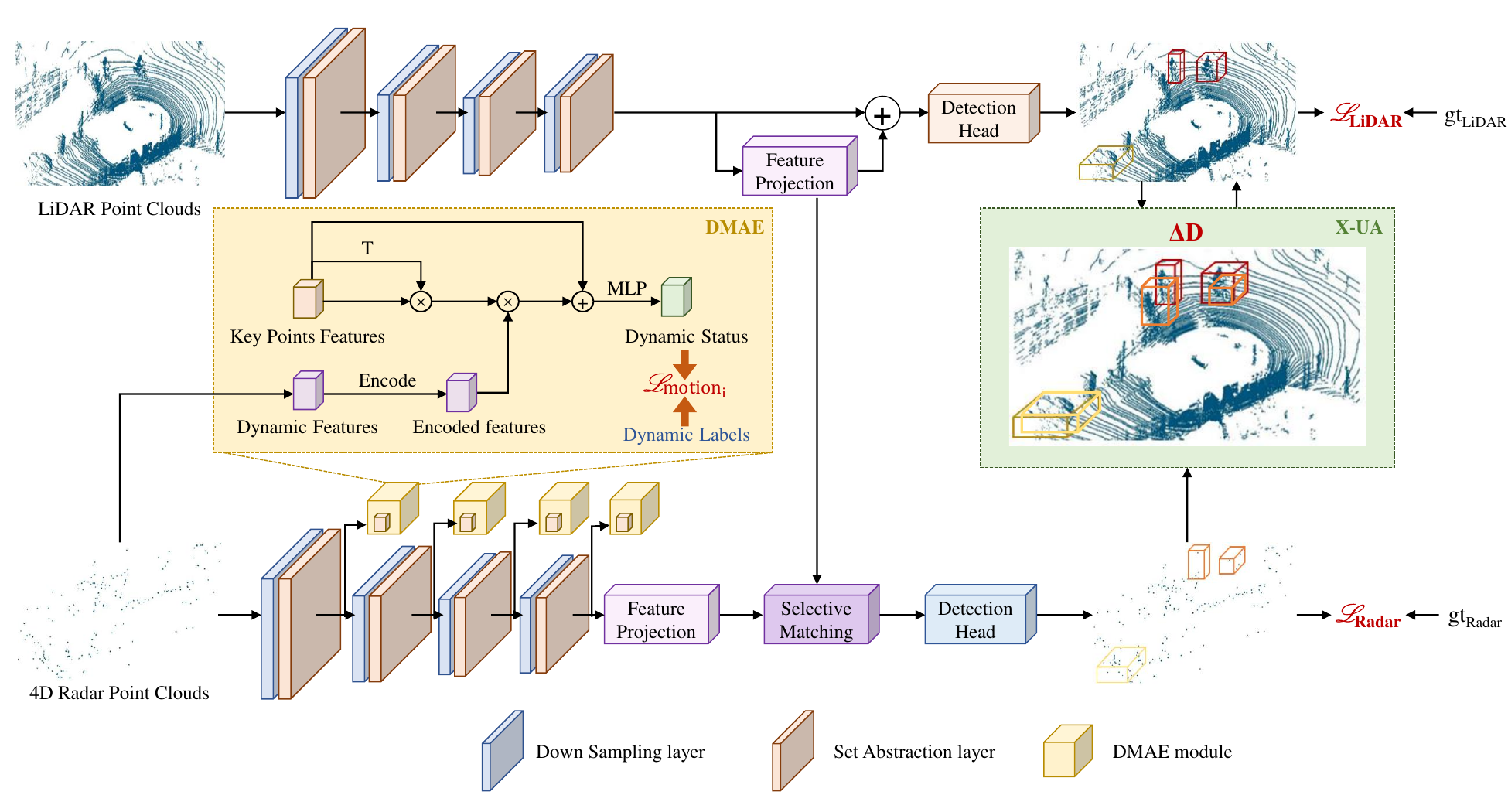} %
    \vspace{-3mm}
    \caption{The overall structure of our ELMAR. The four DMAE modules utilize the single-frame radar scan to predict the motion status of potential objects. The X-UA reduces the misalignment between two predictions.}
    \label{fig: overall structure}
    \vspace{-2mm}
\end{figure*}

The overall structure of our ELMAR is illustrated in Fig. \ref{fig: overall structure}.  
First, the raw LiDAR and 4D radar point clouds are fed into their respective feature extraction encoders. Each encoder consists of four groups of Down-Sampling (DS) and Set Abstraction (SA) layers \cite{qi2017pointnet++} to extract key points features. Each key point in the last layer will generate an bounding box through detection head.

In the 4D radar branch, the key points features from each SA layer are weighted by the measured radar dynamic features and further refined through the DMAE module to implicitly encode the objects' motion status through the motion-aware loss.  
After feature extraction, LiDAR and 4D radar features are projected into a shared latent space and aligned using Feature Projection and Selective Matching, as proposed in \cite{deng2024robust}. Subsequently, two detection heads generate the initial predictions for each modality.  
To address differences in point density and inherent characteristics between LiDAR and 4D radar, the X-UA module further refines the LiDAR-based predictions by incorporating 4D radar predictions and estimating their uncertainties. This approach mitigates potential performance degradation from misaligned predictions.

\subsection{Dynamic Motion-Aware Encoding module}

\begin{figure}[h]
    \centering
    \vspace{-1mm}
    \begin{subfigure}{0.495\linewidth}
        \includegraphics[width=\linewidth]{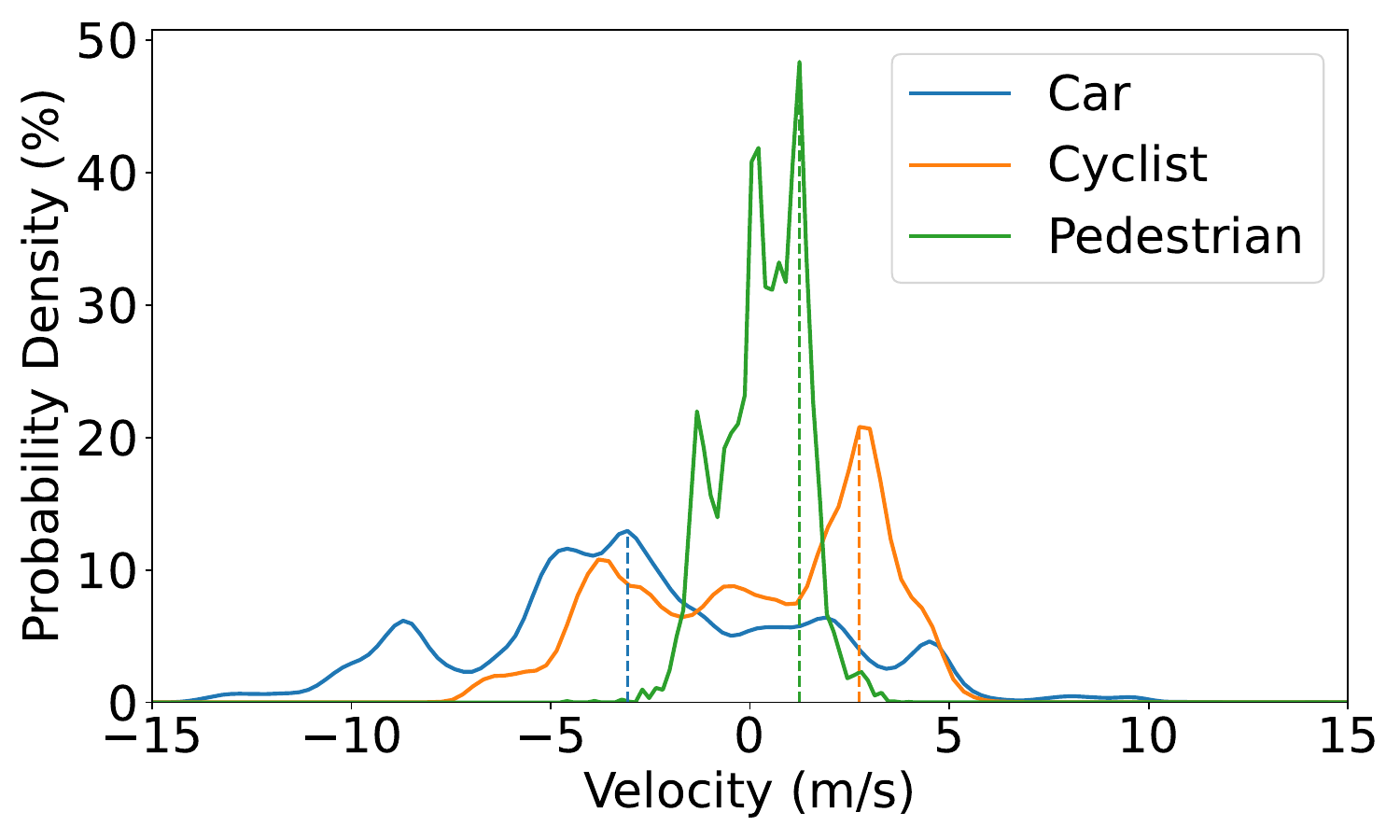}
        \caption{}
        \label{fig:move_objects}
    \end{subfigure}
    \begin{subfigure}{0.49\linewidth}
        \includegraphics[width=\linewidth]{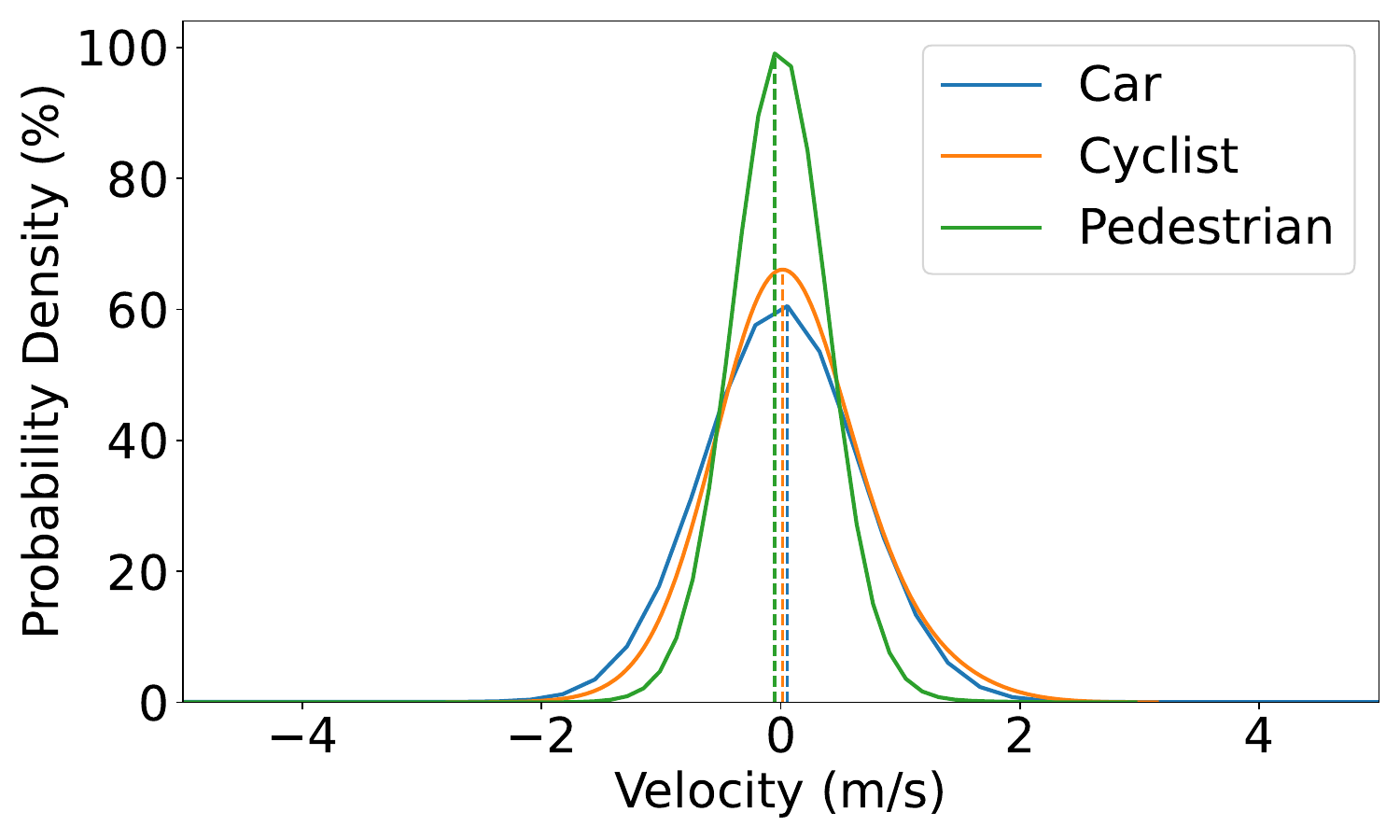}
        \caption{}
        \label{fig:static_objects}
    \end{subfigure}
    \caption{Point velocity distribution of cars, cyclists, and pedestrians in the VoD dataset \cite{palffy2022multi}. (a) shows the absolute velocity distribution in moving objects. (b) shows the absolute velocity distribution in static objects.}
    \label{fig: speed}
    \vspace{-3mm}
\end{figure}

Motion status can provide valuable supplementary information for improving object detection \cite{peng2024mufasa}. For instance, when the background is noisy or the target is partially occluded, dynamic objects can be more easily distinguished based on their moving patterns. However, relying only on measured 4D radar point-wise relative or absolute velocities can lead to errors in determining an object's motion status \cite{10161152}.
As illustrated in Fig. \ref{fig: speed}, some points within the bounding boxes of moving objects still exhibit zero velocity (Fig. \ref{fig:move_objects}), while some points within stationary objects' bounding boxes have nonzero absolute velocity (Fig. \ref{fig:static_objects}). Therefore, using a predefined threshold on points' velocity values to intuitively identify objects’ motion status can lead to substantial inaccuracies and degrade object detection performance. To address this issue, we introduce the DMAE module to encode the objects' dynamic information.
Instead of relying on individual point velocities, our approach constrains the motion status of the entire object through a loss function, allowing the model to implicitly learn more accurate dynamic information of objects, further enhancing detection performance.

In the DMAE module, as shown in Fig. \ref{fig: overall structure},  through the \( i \)-th DS and SA layers (\( i = 1, 2, 3, 4 \)), we extract single-frame 4D radar key points features \( f_i \in \mathbb{R}^{M_i \times N_i} \), where \( M_i \) denotes the number of sampled points, and \( N_i \) represents the feature dimension in the \( i \)-th layer.
However, due to the inherent sparsity of 4D radar point clouds, the extracted key points features \( f_i \) lack detailed spatial information.
Therefore, we incorporate dynamic features to enhance the 4D radar feature representation. Specifically, the measured relative and absolute velocity \( v \in \mathbb{R}^{M_i \times 2} \) are used as dynamic features to weight the key points features \( f_i \) by attention mechanism. The weighted features are then processed through a Multi-Layer Perceptron (MLP) to predict the dynamic status. The predicted dynamic status \( \hat{y}_i \) is formulated as:
\begin{equation}
\hat{y}_i = \text{MLP} \left( \left( (f_i^\top \otimes f_i) \otimes \text{Encode}(v) \right) + f_i \right).
\end{equation}
The predicted dynamic status is further used to calculate the motion-aware loss.

The sub motion-aware loss \( \mathcal{L}_{\text{motion}_i} \) in the \( i \)-th layer is defined as:
\begin{align}
\mathcal{L}_{\text{motion}_i} = -\frac{1}{M_i} \sum^{M_i}_{j=1} \big[ 
& \alpha y^j (1 - \hat{y}^j)^\gamma \log (\hat{y}^j) \nonumber \\
& \hspace{-2em} + (1 - \alpha) (1 - y^j) (\hat{y}^j)^\gamma \log (1 - \hat{y}^j) \big],
\end{align}

where \( y^j \) is the ground truth dynamic status of the \( j \)-th point, determined by the motion status of the entire object. Points within moving objects are labeled as \( y^j = 1 \), while points within static objects are \( y^j = 0 \).
The weighting factor \( \alpha \) balances the static and dynamic points, while the focusing parameter \( \gamma \) from focal loss reduces the relative influence of well-classified points.
Finally, the motion-aware loss \( \mathcal{L}_{\text{Motion}} \) is computed as the average loss across all four layers:
\begin{equation}
\mathcal{L}_{\text{Motion}} = \frac{1}{4} \sum_{i=1}^{4} \mathcal{L}_{\text{motion}_i}.
\end{equation}
The DMAE module encodes dynamic information of 4D radar point clouds, allowing the network to learn the object motion status implicitly. With a single-frame radar scan, it eliminates the need for sequential frame accumulation to infer motion status, avoiding the long-tail problem and high computation cost.

\subsection{Cross-Modal Uncertainty Alignment module}

After the feature encoding, the detection heads generate initial bounding box predictions for both LiDAR and 4D radar branches. However, LiDAR's sensitivity to environments can lead to increased uncertainty in its predictions. In contrast, the 4D radar provides better robustness to noise but may suffer from missed detections due to its sparse point distribution.  
Therefore, inspired by the unsupervised method \cite{zhang2024harnessing}, 
we explore the bounding box distribution of different modalities and design a cross-modal uncertainty alignment module to learn the consistency between two modalities and refine the final LiDAR detection.
\begin{figure}[h]
\vspace{-2mm}
    \centering
    \includegraphics[width=1\linewidth]{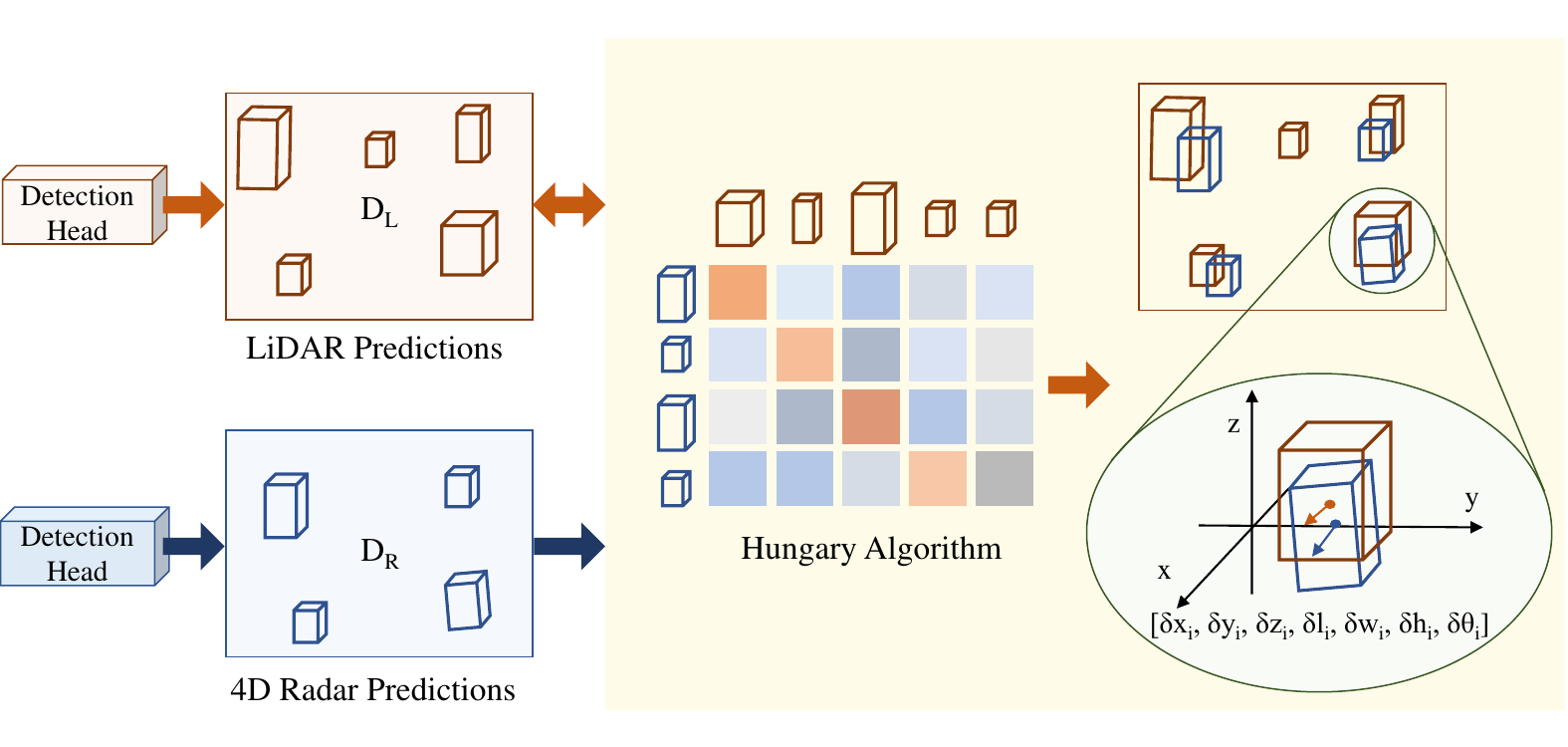}
    \vspace{-5mm}
    \caption{The Structure of U-XA module.}
    \label{fig: the second idea}
    \vspace{-5mm}
\end{figure}

As illustrated in Fig. \ref{fig: the second idea}, \( D_L \in \mathbb{R}^{m \times 7} \) and \( D_R \in \mathbb{R}^{n \times 7} \) represent the initial bounding box predictions from LiDAR and 4D radar, respectively:
\begin{equation}
D_L = \{ (x_L^i, y_L^i, z_L^i, l_L^i, w_L^i, h_L^i, \theta_L^i) \}_{i=1}^{m},
\end{equation}
\begin{equation}
D_R = \{ (x_R^j, y_R^j, z_R^j, l_R^j, w_R^j, h_R^j, \theta_R^j) \}_{j=1}^{n}.
\end{equation}
Here, \( m \) and \( n \) denote the number of detected objects in each modality. Each bounding box is characterized by seven parameters: position (\( x, y, z \)), size (\( l, w, h \)), and orientation (\(\theta\)).
Due to the sparsity gap between LiDAR and 4D radar point clouds, the initial bounding boxes generated by the key points through detection heads can be misaligned. To establish a one-to-one correspondence matching, we use the Hungarian algorithm \cite{kuhn1955hungarian} based on the Euclidean distance between bounding box centers.
The number of matched bounding boxes is defined as \( k = \min(m, n) \).
The aligned bounding box sets, \( \tilde{D}_L \) and \( \tilde{D}_R \), are both with dimensions \( \mathbb{R}^{k \times 7} \). 
After alignment, the difference between the corresponding bounding box sets is computed as:
\begin{align}
\Delta D &= \tilde{D}_L - \tilde{D}_R \nonumber \\
&=
\begin{bmatrix} 
\delta x_1 & \delta y_1 & \delta z_1 & \delta l_1 & \delta w_1 & \delta h_1 & \delta \theta_1 \\
\delta x_2 & \delta y_2 & \delta z_2 & \delta l_2 & \delta w_2 & \delta h_2 & \delta \theta_2 \\
\vdots & \vdots & \vdots & \vdots & \vdots & \vdots & \vdots \\
\delta x_k & \delta y_k & \delta z_k & \delta l_k & \delta w_k & \delta h_k & \delta \theta_k
\end{bmatrix} \in \mathbb{R}^{k \times 7}.
\end{align}
\( \Delta D \) represents the disagreement between the LiDAR and 4D radar predictions in each attribute. A higher \( \Delta D \) indicates greater inconsistency, assigning lower confidence in the final prediction. Therefore, we use \( \Delta D \) as the estimated fine-grained uncertainty.

The original loss function for matched LiDAR predictions is defined as:
\begin{equation}
\mathcal{L}_{LiDAR}^{\text{}} = \mathcal{L}_{L}^{\text{reg}} + \mathcal{L}_{L}^{\text{cls}},
\end{equation}
where \( \mathcal{L}_{L}^{\text{reg}} \) denotes the regression for box localization, and \( \mathcal{L}_{L}^{\text{cls}} \) corresponds to the classification.
To integrate uncertainty into the learning process, we define the uncertainty alignment loss function for the LiDAR branch as follows:
\begin{equation}
\mathcal{L}_{uncertainty}^{} = \mathcal{L}_{LiDAR}^{\text{}} \cdot \exp(-\Delta D) + \lambda \Delta D.
\end{equation} 
The term \( \exp(-\Delta D) \) adaptively scales the bounding box loss \( \mathcal{L}_{Lidar}^{\text{}} \) by reducing its impact on uncertain predictions and allowing the model to focus on more reliable predictions. Meanwhile, the regularization term \( \lambda \Delta D \) prevents excessive uncertainty, ensuring stable training by penalizing large differences between the two modalities.

\section{Experiments}
\label{Experiments}


In this section, we compare our approach with other single-modal and multi-modal 3D detection methods. The models were trained for 80 epochs with a batch size 8 on two NVIDIA Tesla A30 GPUs. Our implementation is based on the OpenPCDet library \cite{od2020openpcdet}.

\subsection{Dataset and Metrics}

Our method is evaluated on the VoD dataset \cite{palffy2022multi}, which comprises 8,600 frames of camera, LiDAR, and 4D radar data, labeled by 3D bounding boxes with the object motion status \footnote{Object motion status label: https://github.com/tudelft-iv/view-of-delft-dataset/issues/53}. Due to the unavailable test server, we conduct evaluations on the validation set \cite{10268601}. Detection performance is measured using Average Precision (AP) for each class and mean Average Precision (mAP) across all classes. The Intersection over Union (IoU) thresholds are set as 50\% for cars, 25\% for cyclists, and 25\% for pedestrians. Results are evaluated in the entire scene and the driving corridor.

\subsection{Main Results}

\setlength\tabcolsep{13pt}
\begin{table*}[ht]
\centering
\caption{Comparative AP results on VoD val. set. The values are in \%. The best results are bold and the second best are marked with underlines.}
\label{tab: main result vod}
\begin{tabular}{c|c|cccc|cccc}
\hline
\multirow{2}{*}{Methods} &\multirow{2}{*}{Modality} & \multicolumn{4}{c|}{All Area}   & \multicolumn{4}{c}{Driving Corridor}\\
\cline{3-10}
 &  &Car & Ped. &Cyc. & mAP & Car & Ped. & Cyc. &  mAP\\
\hline
PointPillars\cite{lang2019pointpillars} &R &37.92   &31.24  &65.66  &44.94  &71.41  &42.27  &87.68  &67.12 \\
MVFAN\cite{yan2023mvfan} &R &38.12 &30.96  &66.17  &45.08  &71.45 &40.21 &86.63 &66.10 \\
PV-RCNN\cite{shi2020pv} &R &41.65   &38.82  &58.36  &46.28  &72.00  &43.53  &78.32  &64.62 \\
SMURF\cite{10274127} &R &42.31 &39.09  &71.50  &50.97  &71.74  &50.54  &86.87  &69.72 \\
MUFASA\cite{peng2024mufasa} &R &43.10  &38.97 &68.65 &50.24 &72.50  &50.28 &88.51  &70.43\\
\cdashline{1-10}
BEVFusion\cite{10160968}  &R+C &37.85 &40.96 &68.95 &49.25 &70.21 &45.86  &89.48  &68.52 \\
RCFusion\cite{zheng2023rcfusion}  &R+C &41.70 &38.95 &68.31 &49.65 &71.87 &47.50  &88.33  &69.23 \\
RCBEVDet\cite{lin2024rcbevdet} &R+C &40.63 &38.86  &70.48  &49.99  &72.48  &49.89  &87.01  &69.80\\
LXL\cite{xiong2023lxl} &R+C &42.33 &49.48 &77.12 &56.31 &72.18 &58.30 &88.31 &72.93 \\
LXLv2\cite{ding2025radarocc} &R+C &47.81 &49.30 &77.15 &58.09 &- &- &- &- \\
SGDet3D\cite{10783046} &R+C &53.16 &49.98 &76.11 &59.75 &81.13 &60.91 &90.22 &77.42 \\
\cdashline{1-10}
PointPillars \cite{lang2019pointpillars} &L &65.55  &55.71  &72.96  &64.74  &81.10  &67.92  &88.96  &79.33 \\
LXL-Pointpillars \cite{xiong2023lxl} &L &66.60 &56.10 &75.10 &65.90 &-  &-  &-  &- \\
\cdashline{1-10}
InterFusion\cite{wang2022interfusion} &R+L &67.50 &63.21  &78.79 &69.83 &88.11  &74.80  &87.50  &83.47\\
L4DR\cite{huang2024l4dr} &R+L &69.10 &66.20 &\textbf{82.80} &72.70  &90.80 &76.10 &\textbf{95.50} & \underline{87.47}\\
MutualForce\cite{peng2025mutualforce} &R+L &71.67 &66.26 &77.35 &71.76  &\textbf{92.31} &76.79&89.97 &86.36  \\
CM-FA\cite{deng2024robust} &R+L &\textbf{77.52} &\underline{67.99}  &75.97  &\underline{73.83} &90.91  &\underline{80.78}  &87.80  &86.50  \\
\cdashline{1-10}
\rowcolor{lightblue}
ELMAR (Ours) & R+L &\underline{76.41}& \textbf{69.34} &\underline{78.91} & \textbf{74.89} &\underline{91.74} & \textbf{81.35} & \underline{93.01} &\textbf{88.70}\\
\hline
\end{tabular}
\\[2pt]
\scriptsize{R, C, and L denote the 4D radar, camera, and LiDAR.}
\vspace{-6mm}
\label{tab: main result vod}
\end{table*}

We conducted extensive evaluations comparing our ELMAR with existing single- and multi-modal 3D object detection approaches. The results are listed in Table \ref{tab: main result vod}.  
From Table \ref{tab: main result vod}, methods incorporating LiDAR achieve significantly better performance than LiDAR-exclusive methods due to the dense 3D spatial information from LiDAR point clouds.  
Among the 4D radar-LiDAR approaches, our method encodes the motion state of objects while maintaining cross-modal consistency through uncertainty modeling. As a result, ELMAR achieves superior performance, with the mAP of 74.89\% across the entire area and 88.70\% within the driving corridor. 
Notably, for small and dynamic objects, which often suffer from sparse point clouds, our approach achieves the best pedestrian detections with 69.34\% AP for the entire scene and 81.35\% AP for the driving corridor. Additionally, for cyclists, ELMAR surpasses CM-FA \cite{deng2024robust} with a 2.94\% and 5.21\% improvement in two evaluation regions.  
Besides, our model delivers a real-time performance at 30.02 FPS.

\begin{figure*}[h]
    \centering
    \begin{subfigure}{0.24\linewidth}
        \includegraphics[width=\linewidth]{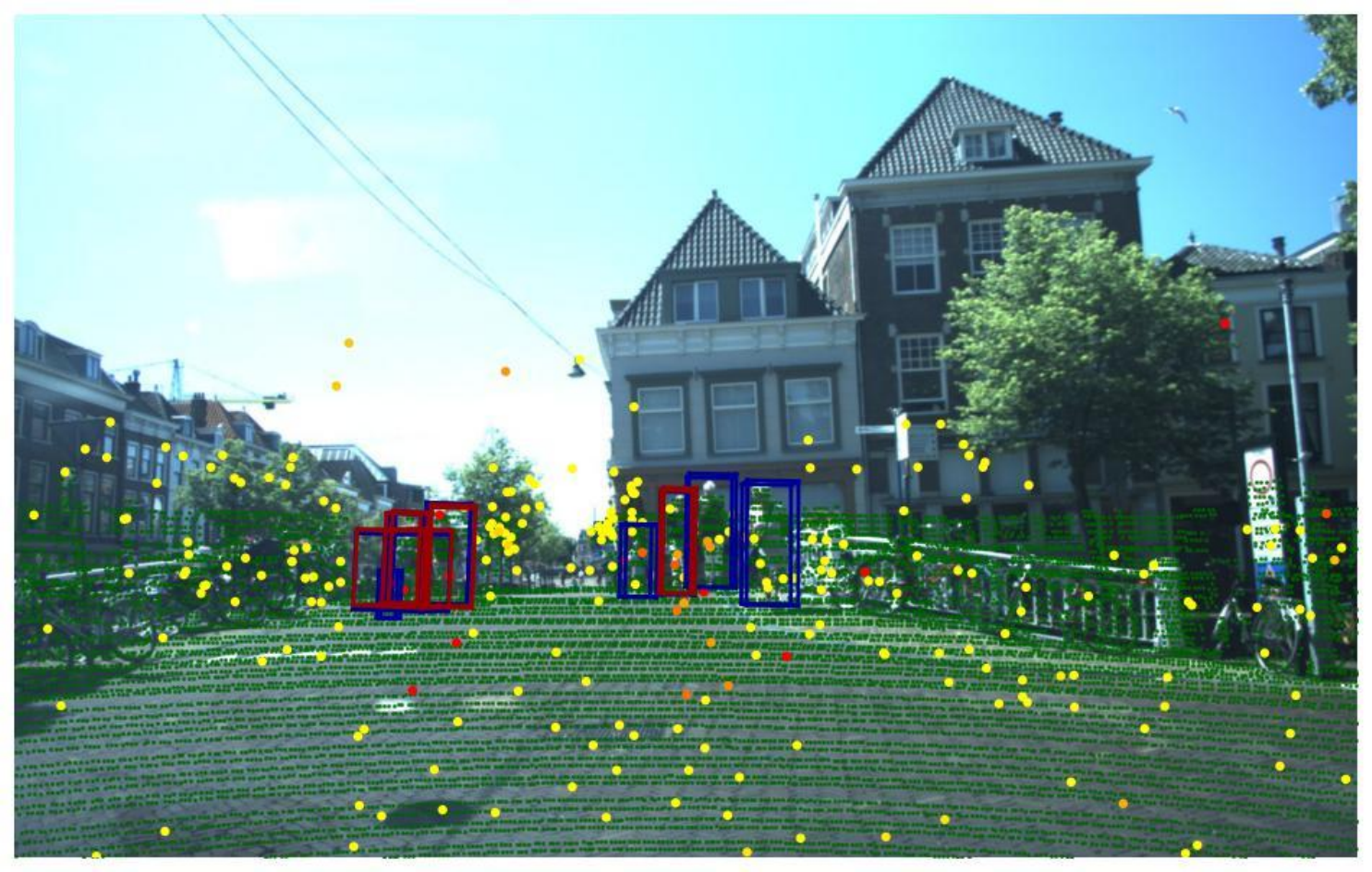}
    \end{subfigure}%
    \begin{subfigure}{0.24\linewidth}
        \includegraphics[width=\linewidth]{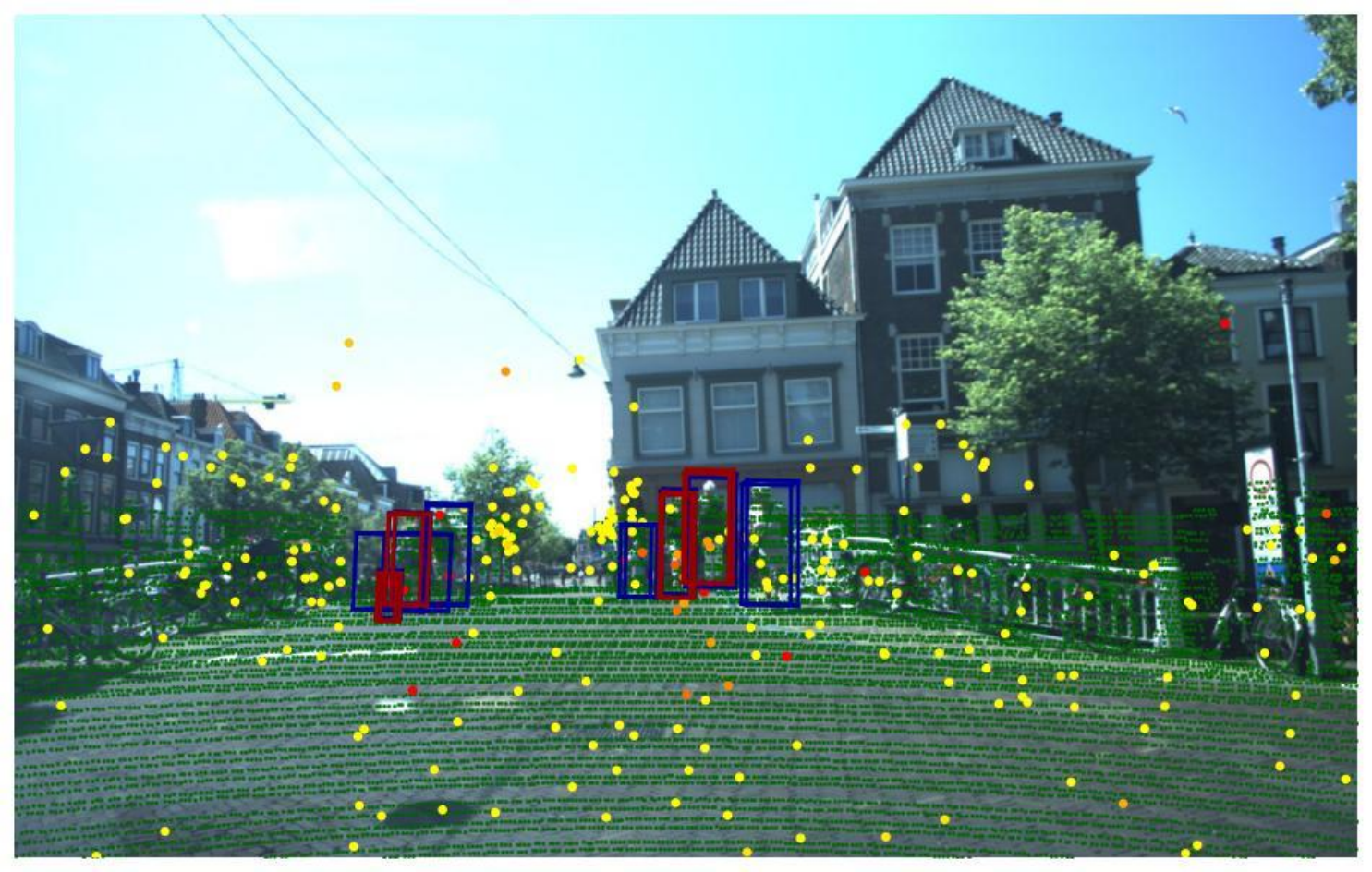}
    \end{subfigure}%
    \begin{subfigure}{0.24\linewidth}
        \includegraphics[width=\linewidth]{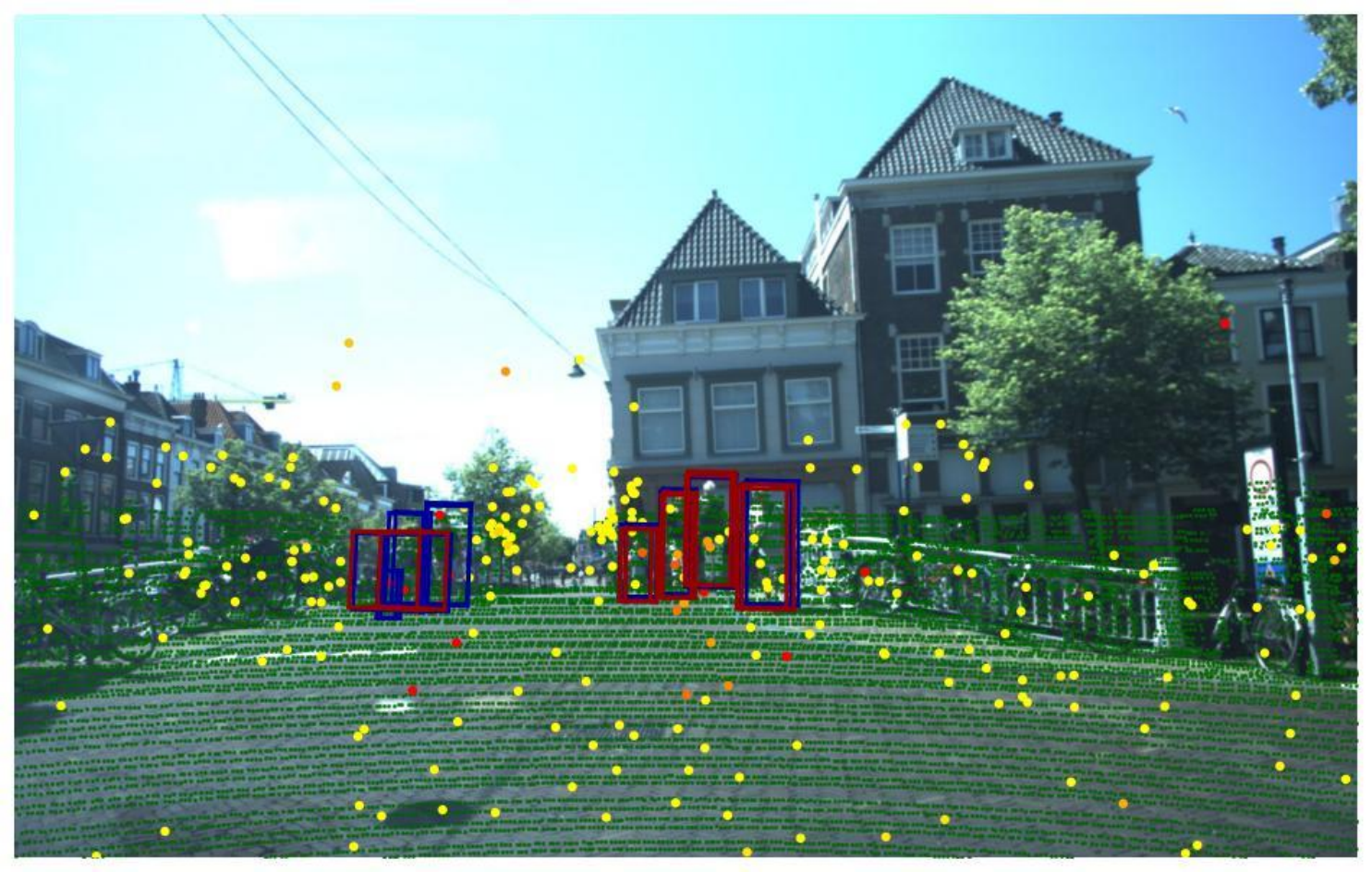}
    \end{subfigure}%
    \begin{subfigure}{0.24\linewidth}
        \includegraphics[width=\linewidth]{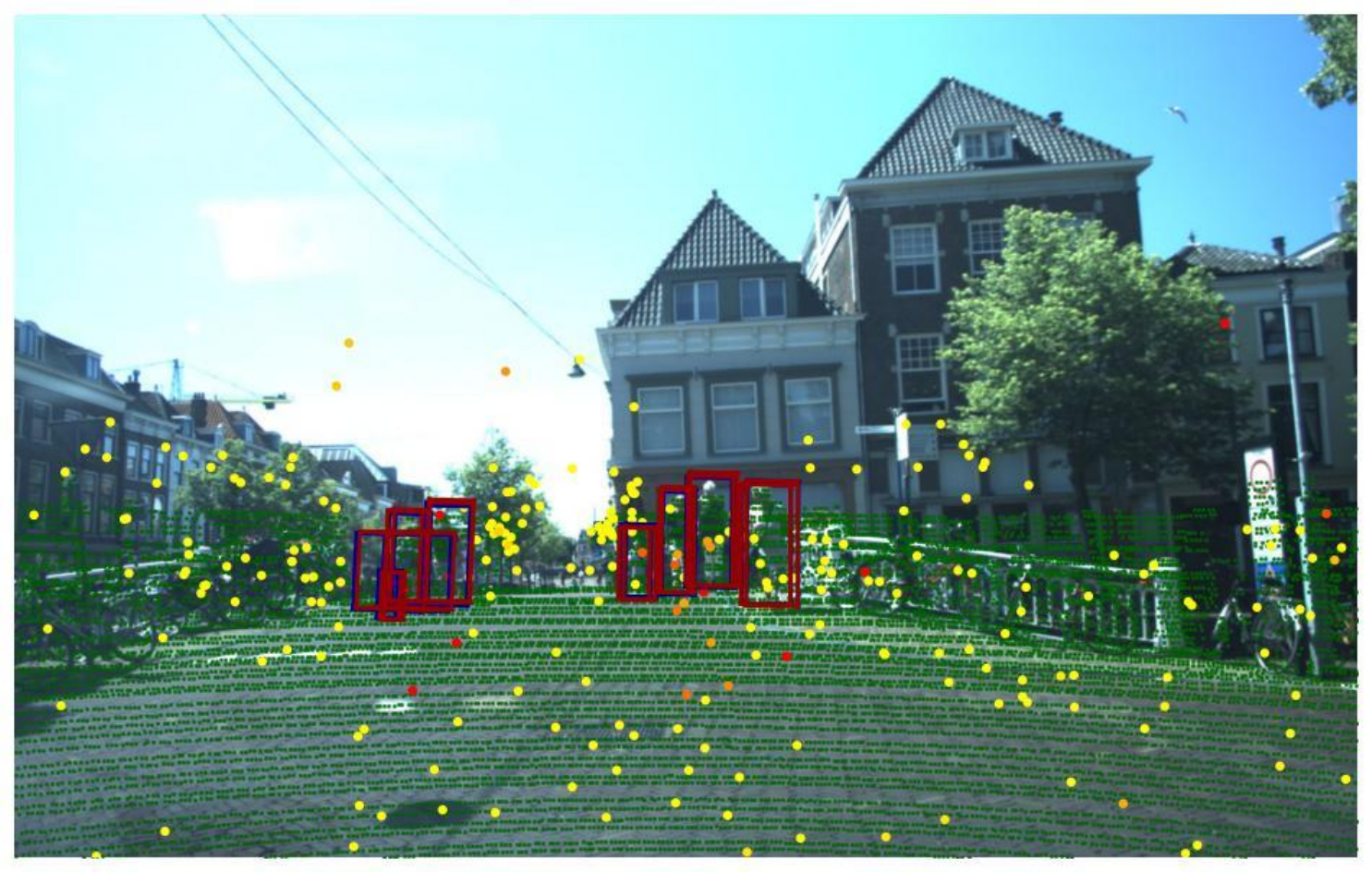}
    \end{subfigure}%

    \begin{subfigure}{0.24\linewidth}
        \includegraphics[width=\linewidth]{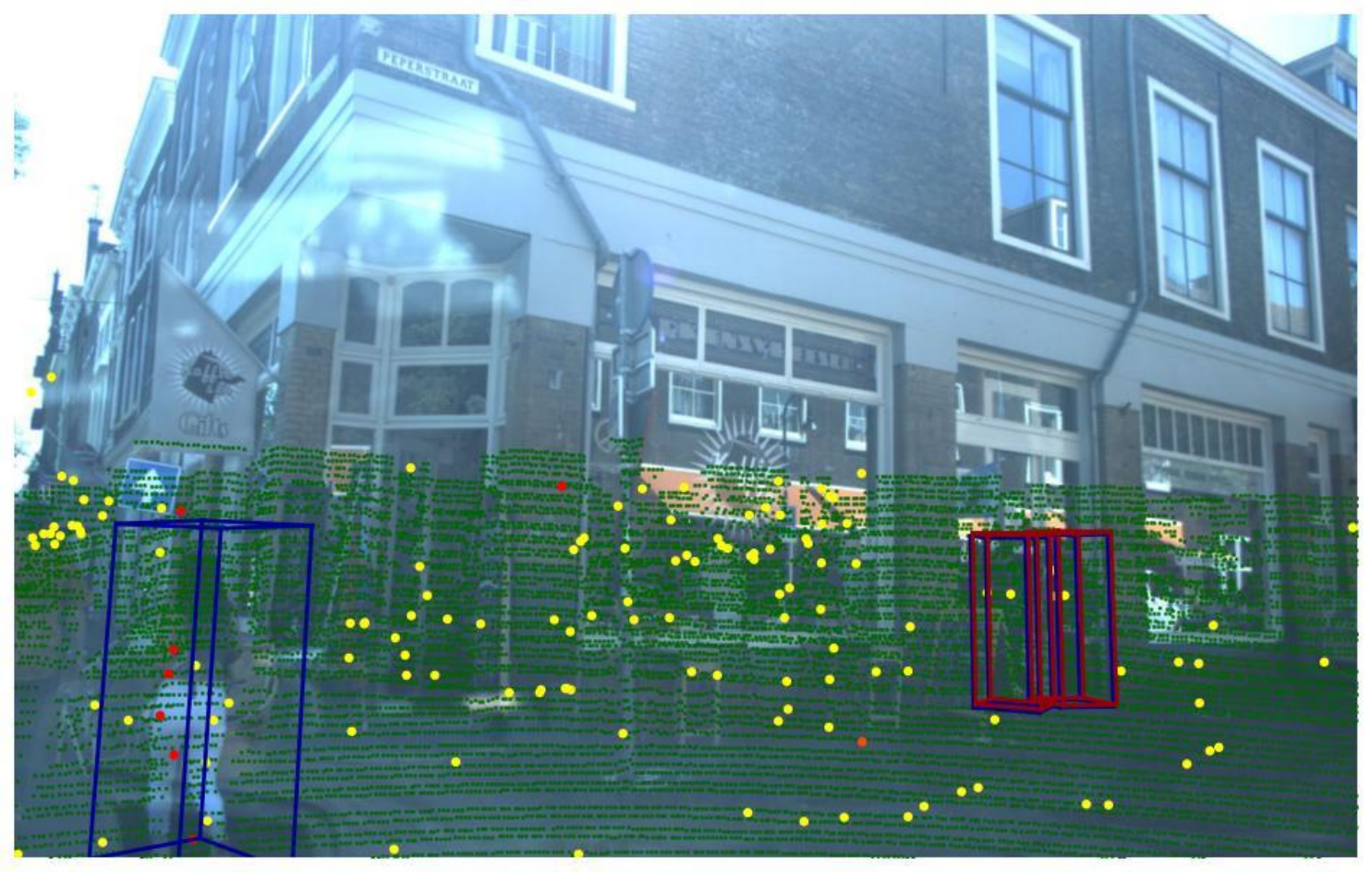}
    \end{subfigure}%
    \begin{subfigure}{0.24\linewidth}
        \includegraphics[width=\linewidth]{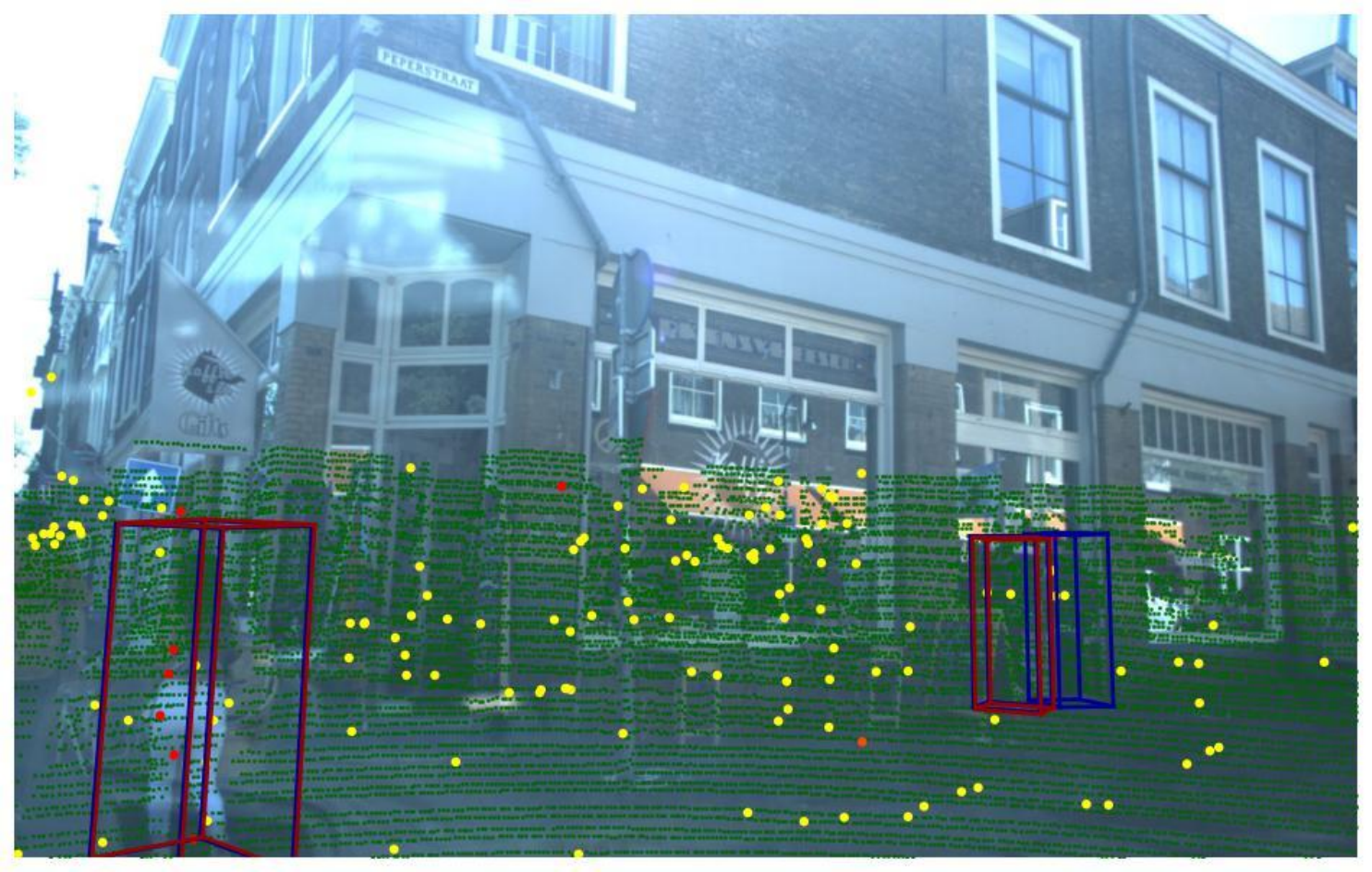}
    \end{subfigure}%
    \begin{subfigure}{0.24\linewidth}
        \includegraphics[width=\linewidth]{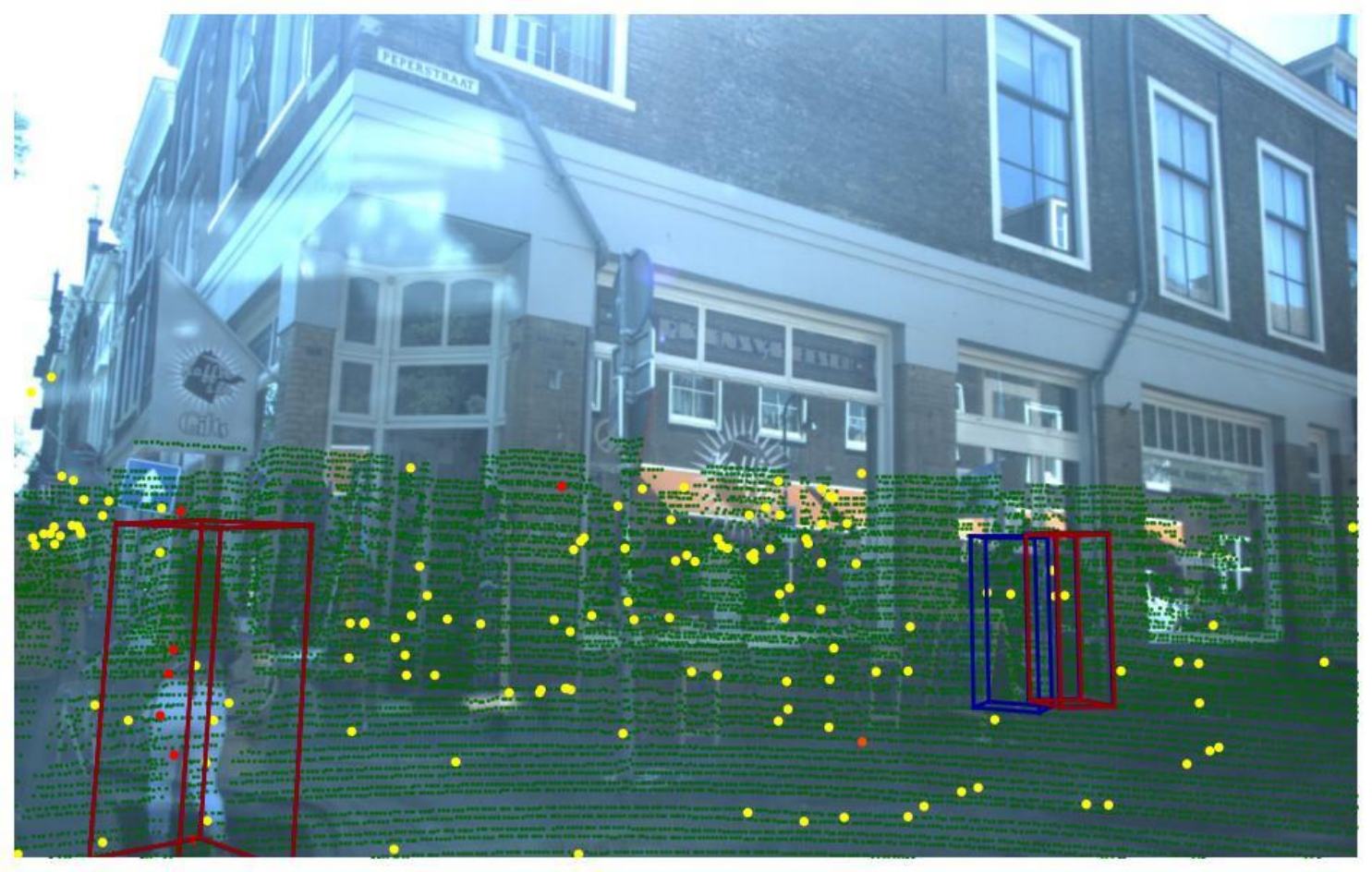}
    \end{subfigure}%
    \begin{subfigure}{0.24\linewidth}
        \includegraphics[width=\linewidth]{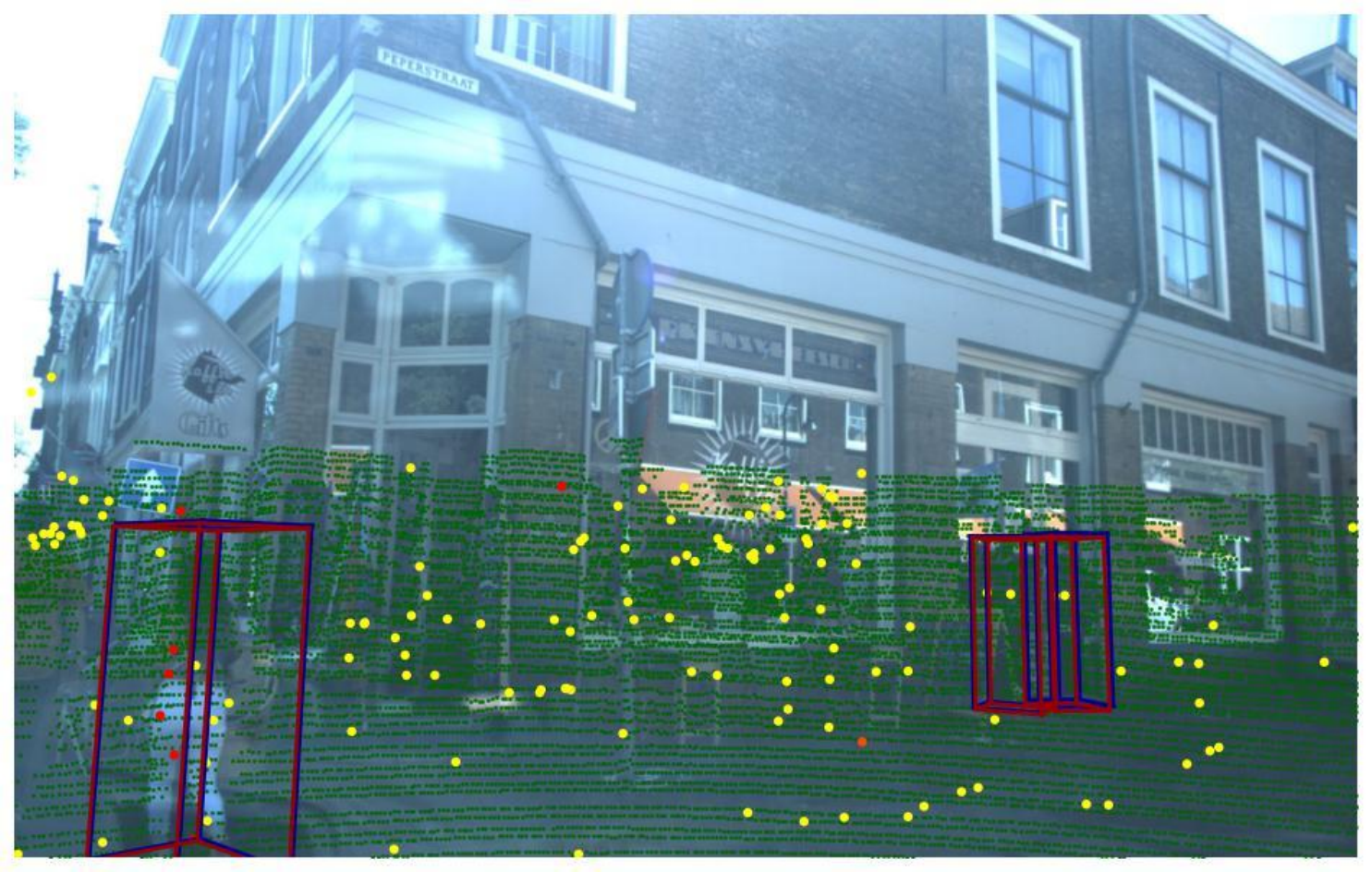}
    \end{subfigure}%

    \vspace {-1mm}
    \makebox[0.24\linewidth]{\textbf{Interfusion \cite{wang2022interfusion}}}%
    \makebox[0.24\linewidth]{\textbf{MutualForce \cite{peng2025mutualforce}}}%
    \makebox[0.24\linewidth]{\textbf{CM-FA} \cite{deng2024robust}}%
    \makebox[0.24\linewidth]{\textbf{ELMAR (Ours)}}
    \caption{Visualization of detections on the VoD dataset \cite{palffy2022multi} across different methods. Ground truth bounding boxes are shown in blue, while predicted bounding boxes are in red. LiDAR points are displayed in green, and 4D radar points are color-coded from yellow to red based on velocity.}
    \label{fig: visualization}
    \vspace{-4mm}
\end{figure*}

We visualize the bounding boxes generated by InterFusion \cite{wang2022interfusion}, MutualForce \cite{peng2025mutualforce}, CM-FA \cite{deng2024robust}, and our proposed method, ELMAR, on the VoD dataset \cite{palffy2022multi}.  
As shown in Fig. \ref{fig: visualization}, ELMAR outperforms existing approaches. Particularly, in detecting small but dynamic objects, ELMAR achieves fewer missed detections. Additionally, our method demonstrates superior effectiveness in handling dense object occlusion.

\subsection{Ablation Study}

We evaluate the effectiveness of proposed modules in Table \ref{tab: modules}. DMAE and X-UA modules are first removed from the framework to establish a baseline.
\setlength\tabcolsep{2.8pt}
\vspace{-1mm}
\begin{table}[h]
\centering
\caption{Ablation study on proposed modules. The values are in \%. The best results are bold.}
\vspace{-2mm}
\label{tab: modules}
\begin{tabular}{c|c|cccc|cccc}
\hline
\multirow{2}{*}{DMAE} & \multirow{2}{*}{X-UA} & \multicolumn{4}{c|}{All Area} & \multicolumn{4}{c}{Driving Corridor} \\
\cline{3-10}
  & & Car & Ped. & Cyc. & mAP & Car & Ped. & Cyc. & mAP \\
\hline
  & &73.51 &63.94 &74.20 &70.55 &90.66 &73.47 &86.50 &83.54 \\
$\checkmark$ & &72.07 &66.61 &77.99 &72.22 &90.79 &76.81 &87.21 &84.94 \\
&$\checkmark$ &\textbf{76.80} &67.50 &77.31 &73.87 &90.47 &78.06 &90.48 &86.34 \\
$\checkmark$ &$\checkmark$ &76.41 &\textbf{69.34} &\textbf{78.91} &\textbf{74.89} &\textbf{91.74} &\textbf{81.35} &\textbf{93.01} &\textbf{88.70} \\
\hline
\end{tabular}
\label{tab: modules}
\vspace{-2mm}
\end{table}

The DMAE module considers the motion status, primarily benefiting the detection of moving objects. Since the VoD dataset \cite{palffy2022multi} was collected in dense urban environments, where most cars are stationary \cite{peng2025mutualforce}, DMAE mainly enhances the pedestrians and cyclists detection by 2.67\% and 3.79\% AP across the entire scene. 
Meanwhile, the X-UA module mitigates the negative impact of radar sparsity on LiDAR detections. By improving cross-modal uncertainty alignment, it achieves a 3.32\% mAP improvement over the baseline.

\textbf{Analysis of Dynamic Motion-Aware Encoding:}
The Dynamic Motion-Aware Encoding with 4D radar velocities is adaptable to various frameworks. We integrate DMAE into the single-modal method MUFASA \cite{peng2024mufasa} and the multi-modal method CM-FA \cite{deng2024robust}. The results in Table \ref{tab: mos} demonstrate its effectiveness across multiple detection frameworks.
\vspace{-2mm}
\setlength\tabcolsep{2.1pt}
\begin{table}[h]
\caption{Comparative AP results for the DMAE on VoD val. set. The values are in \%.}
\label{tab: mos}
\vspace{-2mm}
\begin{tabular}{c|c|cccc|cccc}
\hline
\multirow{2}{*}{Methods} & \multirow{2}{*}{DMAE} & \multicolumn{4}{c|}{All Area}   & \multicolumn{4}{c}{Driving Corridor}\\
\cline{3-10}
 & &Car & Ped. &Cyc. & mAP & Car & Ped. & Cyc. &  mAP\\
\hline
MUFASA& w/o &43.10  &38.97 &68.65 &50.24 &72.50  &50.28 &88.51  &70.43\\
\rowcolor{gray!20}
MUFASA& w &43.59 &40.54 &73.83 &52.65 &73.31 &54.42 &88.98 &73.24 \\
CM-FA& w/o &77.52 &67.99  &75.97  &73.83 &90.91  &80.78 &87.80  &86.50  \\
\rowcolor{gray!20}
CM-FA& w &79.64 &66.74 &77.03 &74.47 &92.84 &81.76 &88.43 &87.68\\
\hline
\end{tabular}
\vspace{-1mm}
\label{tab: mos}
\end{table}

By incorporating motion status to complement static geometric information, the DMAE module enhances original 3D object detection performance. The integration of DMAE leads to 2.41\% mAP improvement for MUFASA \cite{peng2024mufasa} and 0.64\% for CM-FA \cite{deng2024robust}.

\textbf{Analysis of Cross-Modal Uncertainty Alignment:}
\( \lambda \) in the X-UA module controls the penalty strength of high uncertainty predictions. 
A higher \( \lambda \) forces the model to minimize uncertainty aggressively, making it more confident but potentially causing overfitting to noisy data. A lower \( \lambda \) allows more flexibility but risks the model assigning high uncertainty to all predictions. To determine the optimal trade-off, we experimented with different values of \( \lambda \) without the DMAE module, aiming to balance trust in accurate predictions and reduce the influence of unreliable ones. The results of are presented in Table \ref{tab: uncertainty}.
\setlength\tabcolsep{4.6pt}
\vspace{-2mm}
\begin{table}[h]
\centering
\caption{Analysis of \( \lambda \) in the X-UA. The values are in \%. The best results are bold.}\label{tab: uncertainty}
\vspace{-2mm}
\begin{tabular}{c|cccc|cccc}
\hline
\multirow{2}{*}{\( \lambda \)}  & \multicolumn{4}{c|}{All Area}   & \multicolumn{4}{c}{Driving Corridor}\\
\cline{2-9}
 &Car & Ped. & Cyc. & mAP & Car & Ped. & Cyc. & mAP\\
\hline
0.001 &75.41 &64.34 &77.20 &72.32 &90.74 &74.34 &89.21 &84.76\\
0.1 &76.80&\textbf{67.50} &\textbf{77.31} &\textbf{73.87} &90.47 &\textbf{78.06} &\textbf{90.48} &\textbf{86.34}\\
0.5  &\textbf{77.44} &65.10 &75.69 &72.74 &\textbf{90.76} &75.39 &86.57 &84.24 \\
1.0 &77.14 &64.24 &76.53 &72.64 &89.84 &75.79 &86.43 &84.02\\
\hline
\end{tabular}
\vspace{-2mm}
\label{tab: uncertainty}
\end{table}

Our model achieves the highest mAP when \( \lambda \) equals 0.1, showing the best trust in the detections with less uncertainty.



\section{Discussion}
The DMAE module enhances the detection by leveraging objects' velocity. However, in environments where most objects are static, predicting motion states may introduce negative effects on detection. Therefore, our approach is particularly beneficial for dynamic scenes, such as highway. We conducted experiments on the VoD dataset \cite{palffy2022multi}, which provides calibrated object motion dynamic labels compared to other datasets. 
In future work, we will extend our evaluation to new datasets with motion dynamic labels in detection. 

Besides, our current cross-modal uncertainty alignment module is based on bounding boxes. Moving forward, we aim to explore more fine-grained uncertainty modeling strategies, such as feature-based approaches, to further reduce misalignment between modalities.


\section{Conclusion}

LiDAR lacks point-wise dynamic information, while 4D radar point clouds suffer from sparsity. To address these challenges for cross-modal object detection, we propose an enhanced LiDAR detection framework with 4D radar motion
awareness and cross-modal uncertainty.  
During the feature extraction stage, we implement a Dynamic Motion-Aware Encoding module to encode the object motion status and improve 4D radar predictions. In the prediction stage, the uncertainties of initial predictions are estimated to mitigate cross-modal misalignment and refine the final LiDAR predictions. It prevents the sparse 4D radar point clouds from negatively impacting LiDAR spatial sensing capabilities.  
Experiments on the VoD dataset demonstrate the superiority of our approach over existing methods, achieving state-of-the-art mAP. Notably, our method especially excels in detecting small and dynamic objects, improving AP by 1.35\% for pedestrians and 2.94\% for cyclists.





\section*{Acknowledgment}
This research has been conducted as part of the DELPHI project, which is funded by the European Union, under grant agreement No 101104263. 
Views and opinions expressed are those of the author(s) only and do not necessarily reflect those of the European Union or the European Climate, Infrastructure and Environment Executive Agency (CINEA).

\bibliographystyle{IEEEtran} 
\bibliography{thebib} 


\end{document}